\documentclass[11pt]{article}

\usepackage[margin=1in]{geometry}
\usepackage[T1]{fontenc}
\usepackage{times}
\usepackage[hyphens]{url}
\usepackage{graphicx}
\usepackage{amsmath,amssymb}
\usepackage{booktabs}
\usepackage{tabularx}
\usepackage{array}
\usepackage{algorithm}
\usepackage{algorithmic}
\usepackage{natbib}
\usepackage{caption}
\usepackage{xcolor}
\usepackage[colorlinks=true,linkcolor=blue,citecolor=blue,urlcolor=blue]{hyperref}

\newcommand{\method}{Embodied CAD}
\newcommand{\brep}{B-Rep}
\newcommand{\state}{s}
\newcommand{\action}{a}
\newcommand{\traj}{\tau}

\title{Embodied CAD: Solver-Grounded LLM Agents for Parametric B-Rep Assembly Modeling}
\author{
Fumin Liu\thanks{Equal contribution.}\quad
Haoyu Zhou\thanks{Equal contribution.}\quad
Fei Hao\quad
Lin Yang\thanks{Corresponding author.}\\
Nanjing University\\
\texttt{\small 602025710018@smail.nju.edu.cn; haoyu.max@gmail.com}\\
\texttt{\small 6020250201012@smail.nju.edu.cn; linyang@nju.edu.cn}
}
\date{}
\begin{document}

\maketitle

\begin{abstract}
Large language models can write plausible CAD scripts, but reliable industrial CAD modeling requires more than syntactically valid code: every feature, placement, and assembly relation must be accepted by an exact geometric kernel while remaining editable as parametric boundary representation geometry.  We present \method, a closed-loop framework that grounds an LLM agent in a CAD execution environment.  Instead of generating a complete script in one pass, the agent iteratively selects actions from a stratified L0--L4 CAD skill library, resolves them into typed geometric operations, executes them in a CAD backend, and uses solver feedback to plan, repair, and learn.  The framework combines operation-family prediction, deterministic parameter resolution, and solver-derived rewards for supervised warm-up and GRPO-style refinement.  We evaluate \method on multi-step mechanical, industrial-equipment, and mold-oriented assembly tasks using solver-aligned metrics: executable rate, skill accuracy, operation-family accuracy, exact policy accuracy, task completion, and FreeCAD execution success.  The results show that solver-grounded planning executes all strong-planner workflows in the current benchmark, while learned controllers reach high executable rates and expose the remaining gap between valid tool calls and exact long-horizon policy prediction.
\end{abstract}

\section{Introduction}

Large language models (LLMs) have made rapid progress in program synthesis, tool use, and interactive software automation \citep{ouyang2022training,yao2023react,schick2023toolformer,madaan2023selfrefine,shinn2023reflexion}.  Computer-aided design (CAD), however, remains a difficult target for autonomous generation.  A useful engineering CAD artifact is not just a visually plausible mesh.  It must be represented as editable parametric \brep{} geometry, respect dimensional and topological constraints, preserve feature dependencies, and compose multiple parts through precise assembly relations.

This gap is especially visible in long assembly workflows.  An LLM may emit syntactically valid CadQuery or FreeCAD code while placing a cutter in the wrong coordinate frame, selecting a stale local face, duplicating repeated holes with an incorrect offset, or stacking components at the origin.  These errors are often invisible at the token level but fatal at the CAD-kernel level.  Existing sequence, sketch, and text-to-CAD systems have shown strong progress on part-level modeling and command generation \citep{wu2021deepcad,xu2022skexgen,khan2024text2cad,doris2025cadcoder,guan2025cadcoder}, yet industrial assemblies require a different form of grounding: the model must repeatedly act in, inspect, and recover from an exact geometric environment.

\method{} addresses this problem by treating CAD generation as closed-loop embodied tool use.  The agent receives a natural-language specification and a structured parameter set, chooses the next CAD skill, executes it through a deterministic backend, observes solver feedback, and continues until the assembly is complete.  The framework follows the loop
\begin{center}
Intent $\rightarrow$ Skill Planning $\rightarrow$ CAD Execution $\rightarrow$ Solver Feedback $\rightarrow$ Reflection / Policy Update.
\end{center}
The key design choice is to constrain the model to semantically meaningful CAD skills while leaving fragile geometric bookkeeping to deterministic resolvers and the CAD solver.

The paper makes four contributions.  First, we formulate autonomous parametric assembly modeling as a solver-grounded skill-trajectory problem rather than open-loop script generation.  Second, we design an L0--L4 CAD skill hierarchy that spans workspace operations, primitive construction, feature machining, spatial assembly, and domain-level macros.  Third, we introduce operation-family resolution and solver-derived feedback as a bridge between LLM planning and exact \brep{} execution.  Fourth, we evaluate the framework on multi-category industrial assembly tasks with solver-aligned metrics and report both quantitative controller performance and qualitative construction traces.

\begin{figure}[t]
\centering
\includegraphics[width=0.96\textwidth]{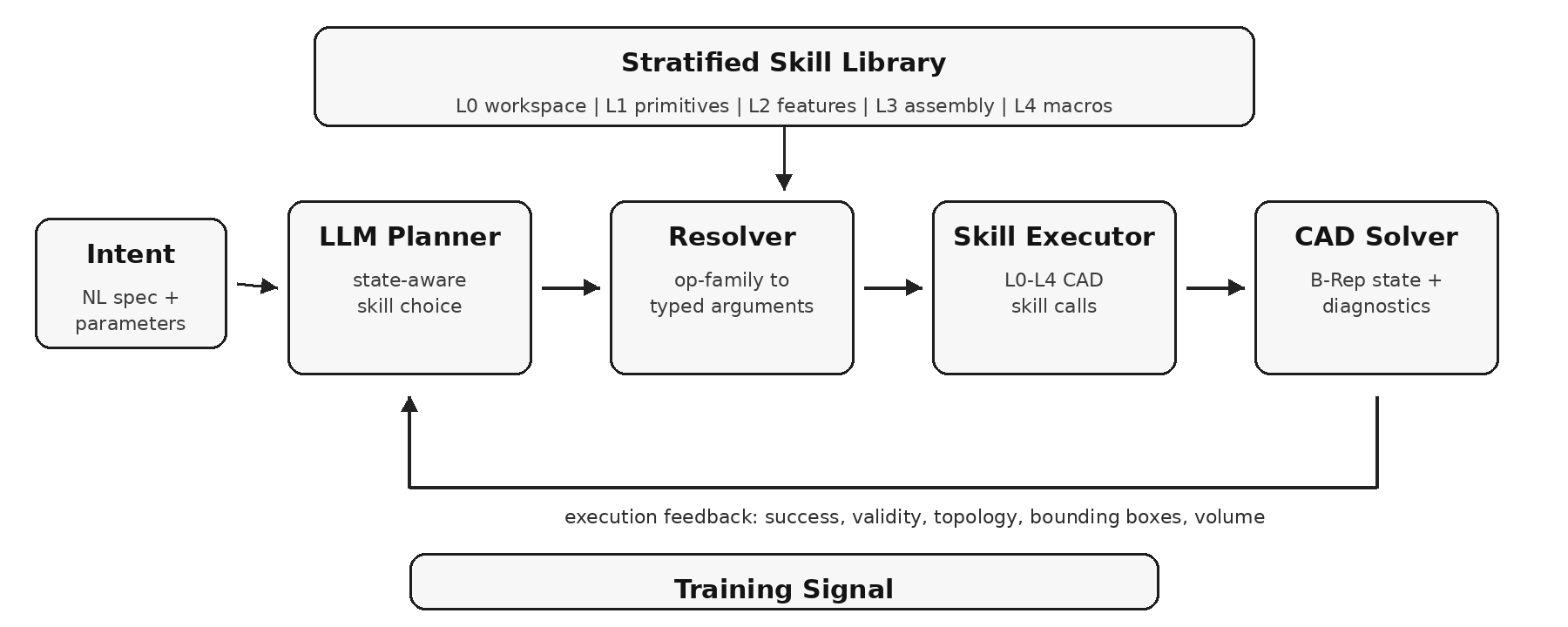}
\caption{Overview of \method. The LLM planner predicts skill-level actions or operation families. A deterministic resolver instantiates typed arguments, the skill executor calls the CAD backend, and the solver returns execution diagnostics. The same feedback supports online repair and training rewards.}
\label{fig:architecture}
\end{figure}

\begin{figure}[t]
\centering
\includegraphics[width=\textwidth]{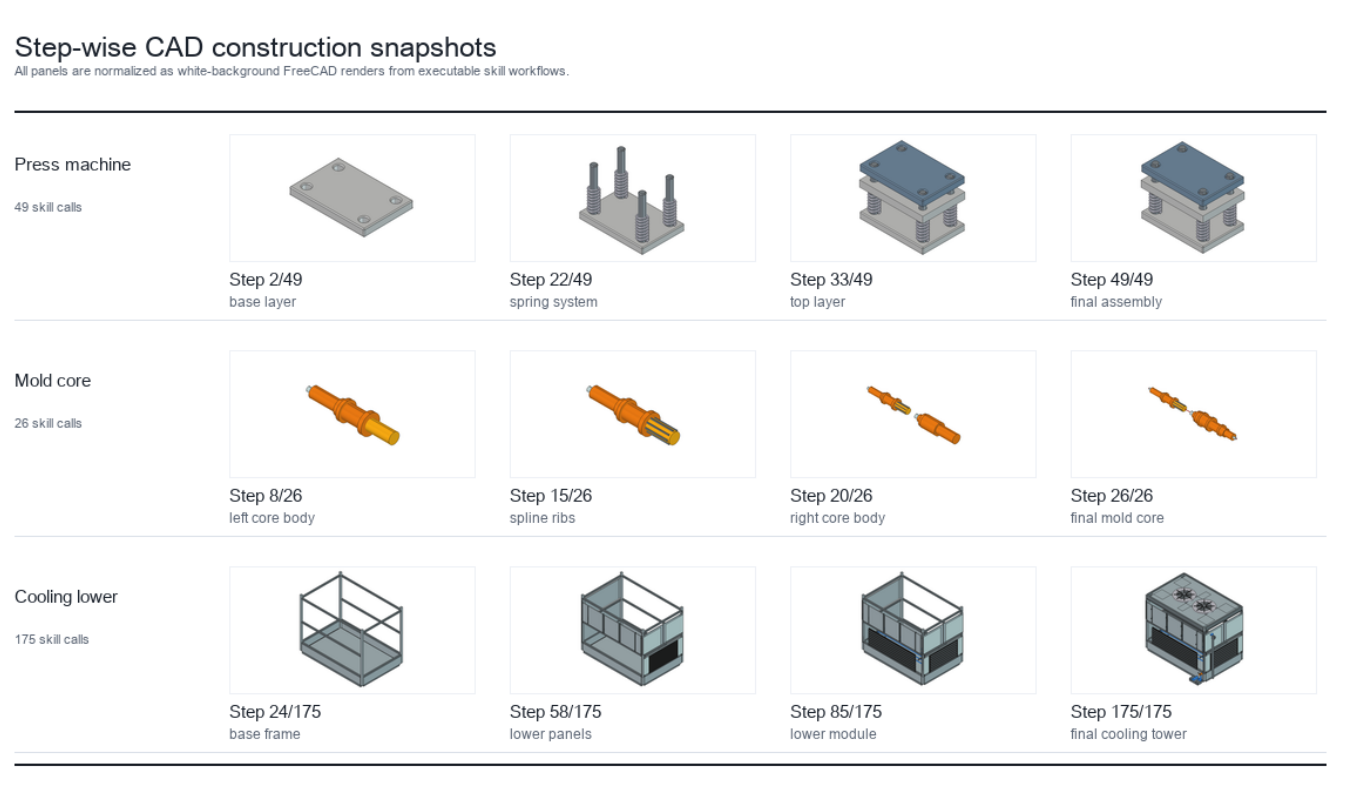}
\caption{Step-wise PhyClaw construction snapshots for three representative strong-planner workflows: a press machine, a mold-core insert, and a cooling-tower workflow whose early checkpoints emphasize the lower module before the final 175-step assembly. Each row uses the same number of solver-checked stages to show base geometry, repeated feature construction, and the final editable state.}
\label{fig:phyclaw-stepwise-snapshots}
\end{figure}

\begin{figure}[t]
\centering
\includegraphics[width=\textwidth]{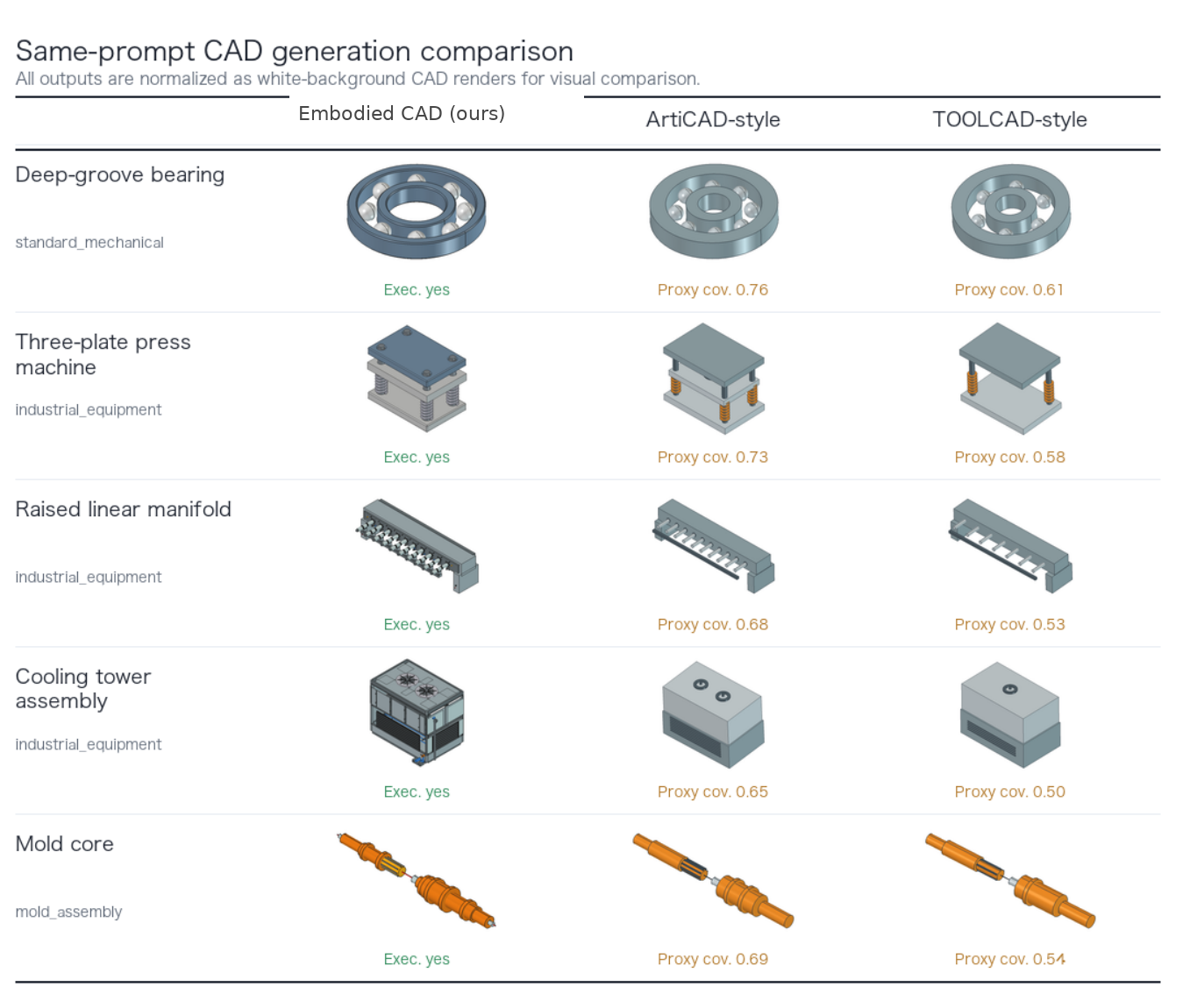}
\caption{FreeCAD-normalized same-prompt qualitative comparison. PhyClaw outputs are executable skill workflows; ArtiCAD-style and TOOLCAD-style columns are paper-inspired proxy baselines rendered through the same FreeCAD viewer for visual style consistency. The proxy columns compare visual feature coverage and are not official executable reproductions.}
\label{fig:same-prompt-proxy-comparison}
\end{figure}

\begin{figure}[t]
\centering
\includegraphics[width=\textwidth]{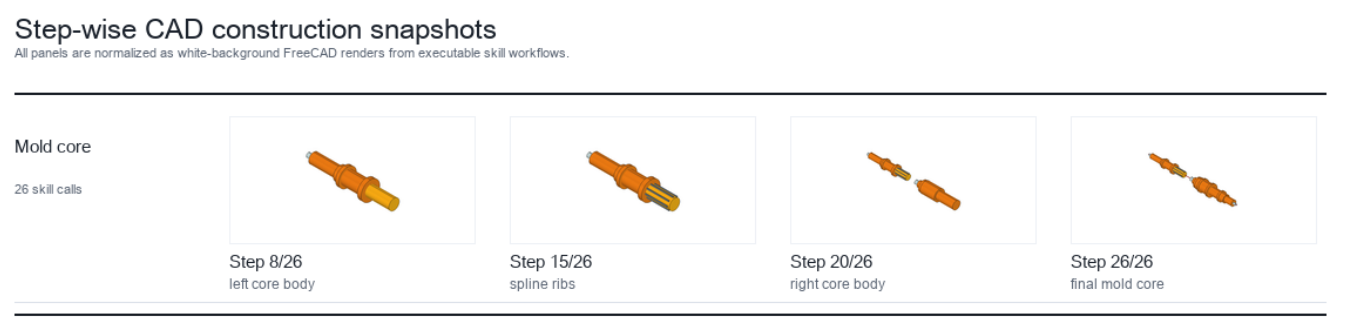}
\caption{Mold-core insert construction snapshots for the current strong-planner POC path. The same parameterized workflow supports fast regeneration after user edits to shaft radii, rib count, and segment layout.}
\label{fig:mold-core-stepwise}
\end{figure}

\section{Related Work}

\textbf{Parametric CAD generation.}
Large CAD datasets and sequence models have made parametric CAD generation a practical learning problem.  ABC provides a large \brep{}-oriented corpus for geometric learning \citep{koch2019abc}; the Fusion 360 Gallery dataset exposes human design histories for programmatic CAD construction \citep{willis2021fusion}.  DeepCAD models CAD commands as a sequence generation problem \citep{wu2021deepcad}, SkexGen improves sequence generation with disentangled codebooks \citep{xu2022skexgen}, and Text2CAD maps natural language to CAD command sequences \citep{khan2024text2cad}.  Recent CAD-code systems further explore vision- and text-conditioned code generation \citep{doris2025cadcoder,guan2025cadcoder}.  In contrast, \method{} does not ask the model to emit an entire CAD program in one pass.  It constrains generation to solver-checked skill trajectories whose intermediate states can be inspected and repaired.

\textbf{B-Rep learning and executable structure.}
Engineering CAD models contain topology, analytic surfaces, curves, feature history, and constraints that are not captured by visual surface similarity alone.  BRepNet performs topological message passing over solid models \citep{lambourne2021brepnet}; UV-Net learns from parametric curve and surface domains \citep{jayaraman2021uvnet}; SolidGen and BRepGen target direct boundary-representation synthesis \citep{jayaraman2022solidgen,xu2024brepgen}.  Programmatic approaches such as CSGNet and ShapeAssembly show the value of explicit executable structure \citep{sharma2018csgnet,jones2020shapeassembly}.  \method{} shares this preference for executable structure but focuses on long, editable assembly workflows in which each intermediate CAD state is validated by a solver.

\textbf{Assembly generation and tool-using agents.}
Assembly modeling adds repeated parts, joint relations, and spatial constraints beyond single-part generation.  JoinABLe studies bottom-up assembly of parametric CAD joints \citep{willis2022joinable}.  Tool-augmented CAD systems, including CAD-Assistant, ArtiCAD, and TOOLCAD, connect foundation models to CAD tools or tool-use training pipelines \citep{mallis2025cadassistant,shui2026articad,gong2026toolcad}.  Our work is closest to these tool-augmented systems, but emphasizes typed skill calls, operation-family resolution, and solver-aligned evaluation rather than visual-only judging.

\section{Problem Formulation}

Given an industrial modeling request $x$ and a parameter set $p$, the goal is to produce an executable trajectory
\begin{equation}
\traj = (\action_1, \ldots, \action_T), \qquad
\action_t = (\ell_t, f_t, \theta_t),
\end{equation}
where $\ell_t$ is the skill level, $f_t$ is a skill or operation family, and $\theta_t$ contains typed arguments.  A CAD kernel $K$ deterministically updates the state:
\begin{equation}
\state_t = K(\state_{t-1}, \action_t).
\end{equation}
A trajectory is solver-feasible if every action executes without kernel error and the final state satisfies task-specific checks such as valid solids, expected components, bounding boxes, feature counts, and assembly completion.  The central challenge is that $\theta_t$ often depends on the current geometric state: a hole center, chamfer target, mirrored instance, or repeated fastener index must be selected consistently with previous operations.

\method{} separates this problem into planning, resolution, and execution.  The LLM predicts high-level skills or operation families; deterministic modules derive geometry-dependent arguments; the CAD backend executes each skill; and the feedback is returned to the planner and to the learning objective.  This separation reduces hallucinated coordinates and makes failure modes observable.

\section{Method}

\subsection{Closed-Loop Architecture}

Figure~\ref{fig:architecture} summarizes the architecture.  The planner observes the user intent, parameters, construction history, and solver feedback.  It selects the next skill or operation family.  The resolver maps reusable operation families, such as \texttt{build\_pin\_body} or \texttt{cut\_hole\_grid}, to concrete instance identifiers and typed parameters.  The executor dispatches the resolved call to the CAD backend.  The solver returns structured feedback, including success or failure, object names, validity checks, volumes, bounding boxes, and topology summaries when available.

\begin{algorithm}[t]
\caption{Closed-loop CAD skill execution}
\label{alg:closedloop}
\begin{algorithmic}[1]
\STATE Initialize CAD state $\state_0$ from intent $x$ and parameters $p$.
\FOR{$t=1,\ldots,T$}
  \STATE Planner proposes skill or operation family $f_t$.
  \STATE Resolver derives typed arguments $\theta_t$ from $\state_{t-1}$.
  \STATE Execute $\action_t=(\ell_t,f_t,\theta_t)$ with CAD kernel $K$.
  \STATE Collect solver feedback $o_t$ from the resulting state $\state_t$.
  \IF{execution fails}
     \STATE revise $f_t$ or $\theta_t$ using $o_t$ and retry within budget.
  \ENDIF
\ENDFOR
\STATE Return the final parametric \brep{} assembly and trajectory log.
\end{algorithmic}
\end{algorithm}

\subsection{Skill Stratification}

A raw CAD scripting space is too large for reliable long-horizon LLM planning.  We therefore organize actions into five levels (Table~\ref{tab:skills}).  The hierarchy keeps the model's choices close to engineering operations while preserving enough expressivity to build multi-part assemblies.

\begin{table}[t]
\centering
\small
\begin{tabularx}{\textwidth}{p{0.08\textwidth}p{0.20\textwidth}X}
\toprule
Level & Role & Representative skills \\
\midrule
L0 & Workspace and inspection & Reset workspace, import references, parse parameters, query handbook entries, render checkpoints, inspect solids and bounding boxes. \\
L1 & Primitive construction & Boxes, cylinders, oriented cylinders, spheres, truncated cones, plates, spline extrusions, profile-based solids, and reusable cutters. \\
L2 & Machining features & Boolean cuts, hole grids, ring cutters, spline cutters, chamfers, fillets, washers, nuts, counterbores, and repeated feature arrays. \\
L3 & Spatial assembly & Move, align, mirror, pattern, mate, group, finalize assembly, and export editable CAD states. \\
L4 & Domain macros & Higher-level modules such as bearings, springs, gears, press plates, cooling-tower modules, mold plates, and mold-core inserts. \\
\bottomrule
\end{tabularx}
\caption{L0--L4 skill hierarchy used by \method.  The hierarchy constrains the action space while keeping operations interpretable and editable.}
\label{tab:skills}
\end{table}

\subsection{Operation Families and Deterministic Resolution}

Industrial assemblies contain repeated structures: balls in bearings, pins in manifolds, screws in mold plates, guide posts in press machines, and ribs on shafts.  Directly predicting instance-specific operation keys entangles semantic intent with fragile indexing.  We instead ask the model to predict operation families.  The controller resolves a family into the next valid instance using construction state, counters, symmetry rules, and derived geometry.  This design preserves the agentic planning role of the LLM while moving coordinate-sensitive computations into deterministic modules.

For example, after the planner predicts a family such as \texttt{place\_guide\_post}, the resolver determines which post is next, retrieves the corresponding plate corner, computes the local axis and clearance, and produces the typed skill call.  The solver then checks whether the post intersects the expected plate stack and whether the resulting solid remains valid.  Failures are returned as structured feedback rather than hidden in a failed final script.

\subsection{Learning from Solver Feedback}

We use solver feedback in two ways.  First, supervised fine-tuning warms up the planner on validated trajectories.  Second, GRPO-style refinement samples multiple candidate actions for the same state and scores them with a structured reward:
\begin{equation}
R = w_f R_{\mathrm{format}} + w_p R_{\mathrm{policy}} + w_e R_{\mathrm{exec}}.
\end{equation}
Here $R_{\mathrm{format}}$ checks whether the model emits a parseable action block, $R_{\mathrm{policy}}$ checks whether the selected skill or operation family matches the expected transition, and $R_{\mathrm{exec}}$ measures whether the resolved action executes successfully in the CAD backend.  The relative reward within each group updates the policy.  We report skipped-update and fallback-SFT rates because a high parse rate alone does not guarantee informative reward variance.

\section{Benchmark and Metrics}

The benchmark is organized around executable CAD-skill trajectories.  Each task provides a natural-language industrial specification and a structured parameter set.  The system must generate a sequence of skill calls that can be resolved and executed by the FreeCAD backend.  The current task families are:

\textbf{Standard mechanical assemblies:} bearing tasks involving coaxial rings, polar ball arrays, repeated separators, editable ring dimensions, and chamfered mechanical details.

\textbf{Industrial equipment assemblies:} press-machine, cooling-tower, and linear-manifold tasks involving layered structures, repeated connectors, springs, guide posts, fans, holes, and pipe-like features.

\textbf{Mold assemblies:} mold-machine and mold-core insert tasks involving plate stacks, guide posts, hole grids, counterbores, grooves, spline ribs, and editable core-shaft layouts.

We evaluate two evidence levels.  The \emph{solver-execution level} measures whether a resolved trajectory executes in the CAD backend.  The \emph{learned-policy level} measures whether an SFT or GRPO controller predicts the correct next skill or operation family.  The distinction is important: deterministic strong-planner trajectories test the capacity of the skill library and resolver, whereas learned-policy metrics measure how much of the planning decision can be delegated to the model.

Metrics are solver-aligned.  \emph{Valid/executable rate} checks parseability or FreeCAD execution success.  \emph{Skill accuracy} checks whether the predicted skill matches the target.  \emph{Operation-family accuracy} checks the reusable family.  \emph{Exact policy accuracy} requires action, skill, and family to be correct for the current state.  \emph{Task completion} evaluates all steps in a full workflow when the workflow log is available.  For GRPO, we additionally report group-valid rate, group-exact rate, skipped-update rate, and fallback-SFT rate.

\begin{table}[t]
\centering
\small
\begin{tabularx}{\textwidth}{p{0.22\textwidth}Xcccc}
\toprule
Policy variant & Evaluation protocol & Valid/Exec. & Skill & Op. family & Exact \\
 & & (\%) & (\%) & (\%) & (\%) \\
\midrule
Strong planner & Best solver-checked workflow per task family & 100.0 & -- & -- & -- \\
Weak-planner SFT & State policy, full validation & 100.0 & 97.8 & 93.1 & 93.1 \\
Weak-planner GRPO & State policy, full validation & 100.0 & 97.8 & 93.2 & 93.2 \\
Skill-family SFT & Direct family prediction, stress aggregate & 98.7 & 83.5 & 79.9 & 76.6 \\
\bottomrule
\end{tabularx}
\caption{Main solver-aligned performance by controller mode.  The strong-planner row reports FreeCAD execution success for deterministic workflows, so learned-action accuracy columns are not applicable.  Weak-planner rows evaluate compact state-action prediction.  The skill-family row aggregates boundary-value, random-parameter, and language-paraphrase stress tests for direct next-family prediction.}
\label{tab:main}
\end{table}

\section{Results}

\subsection{Solver-Aligned Performance}

Table~\ref{tab:main} shows the main quantitative results.  Deterministic strong planning executes all evaluated workflow families in the current benchmark.  The weak-planner SFT policy reaches 93.1\% exact accuracy on the state-policy benchmark, and GRPO refinement gives a small gain to 93.2\%.  The direct skill-family SFT model is more challenging because it must predict the family directly under boundary, random-parameter, and paraphrase stress tests; it reaches 76.6\% exact accuracy in the aggregate.

The GRPO run contains 18,335 candidate groups.  Its group-valid rate is 100.0\%, and its group-exact rate is 73.5\%.  The skipped-update rate is 68.3\%, and the fallback-SFT rate is 10.3\%.  These diagnostics indicate that format validity is not the bottleneck; the main learning challenge is obtaining reward variance that distinguishes plausible but geometrically incorrect next actions.

\begin{table}[t]
\centering
\small
\begin{tabularx}{\textwidth}{p{0.19\textwidth}p{0.24\textwidth}p{0.20\textwidth}p{0.12\textwidth}r}
\toprule
Controller & Evidence scope & Task family & Metric & Value (\%) \\
\midrule
Strong planner & FreeCAD execution & bearing & Exec. & 100.0 \\
Strong planner & FreeCAD execution & press machine & Exec. & 100.0 \\
Strong planner & FreeCAD execution & cooling tower & Exec. & 100.0 \\
Strong planner & FreeCAD execution & linear manifold & Exec. & 100.0 \\
Strong planner & FreeCAD execution & mold core & Exec. & 100.0 \\
Weak-planner SFT & State policy & linear manifold & Exact & 93.1 \\
Weak-planner GRPO & State policy & linear manifold & Exact & 93.2 \\
Skill-family SFT & Direct next-family prediction & bearing & Exact & 86.8 \\
Skill-family SFT & Direct next-family prediction & sprocket shaft & Exact & 74.8 \\
Skill-family SFT & Direct next-family prediction & press machine & Exact & 62.8 \\
Skill-family SFT & Direct next-family prediction & linear manifold & Exact & 79.4 \\
Skill-family SFT & Direct next-family prediction & mold machine & Exact & 76.6 \\
\bottomrule
\end{tabularx}
\caption{Task-level evidence in the current artifact snapshot.  We list only evaluated task-method pairs to avoid implying that missing entries are zero-valued results.}
\label{tab:tasklevel}
\end{table}

\subsection{Task-Level Evidence and Ablations}

Table~\ref{tab:tasklevel} breaks the evidence down by controller and task family.  The strong-planner rows confirm that the skill library and deterministic resolver can generate complete solver-checked assemblies for the evaluated families.  The learned-policy rows show that exact next-action prediction remains task dependent.  Bearing tasks are relatively regular because repeated features follow polar patterns, while press-machine and mold-machine tasks require more mixed feature orders and spatial relations.

The ablation trend supports the design choice behind operation families.  Direct family prediction exposes the model to many similar but index-sensitive decisions, producing lower exact accuracy.  The weak-planner state policy predicts a compact controller action and leaves deterministic geometry and instance resolution to the controller, which improves exact state-policy performance.  GRPO refinement provides a limited gain in the current logs, but the skipped-update rate suggests that future reward shaping should create more informative contrasts among candidate actions.

\subsection{Qualitative Analysis}

Figure~\ref{fig:phyclaw-stepwise-snapshots} visualizes intermediate states rather than only final renders.  This is useful for CAD-agent evaluation because many failures occur before the final view: a guide post can be created at the wrong height, a cutter can remove the wrong plate, or an array can be mirrored around the wrong origin.  Intermediate solver-checked snapshots make these errors diagnosable.

Figure~\ref{fig:same-prompt-proxy-comparison} compares same-prompt outputs against paper-inspired ArtiCAD-style and TOOLCAD-style proxy baselines rendered in the same viewer.  These proxy columns are not official reproductions and should not be interpreted as definitive baseline numbers.  They are included as diagnostic comparisons for visual feature coverage under the same prompts.  The key difference is that \method{} produces executable skill workflows, whereas the proxy outputs are evaluated by visual coverage only.

\begin{table}[t]
\centering
\small
\begin{tabular}{lccc}
\toprule
Method & Visual & Exec. & Coverage \\
\midrule
\method{} best & 1.000 & 1.000 & 1.000 \\
ArtiCAD-style proxy & 1.000 & 0.000 & 0.702 \\
TOOLCAD-style proxy & 1.000 & 0.000 & 0.552 \\
\bottomrule
\end{tabular}
\caption{Same-prompt diagnostic comparison.  Coverage is exact workflow coverage for \method{} and heuristic feature coverage for proxy baselines.}
\label{tab:proxy}
\end{table}

\subsection{Failure Modes}

The remaining failures fall into four recurring categories.  \emph{Family confusion} occurs when the model selects a plausible but wrong next operation, such as a chamfer before all cuts have been applied.  \emph{Index ambiguity} occurs when repeated instances are not grounded to the construction state.  \emph{Coordinate-frame errors} occur when local and global frames are mixed, especially for mirrored or rotated features.  \emph{Reward sparsity} occurs when many sampled candidates are parseable and executable but have identical group rewards.  Operation-family resolution reduces the second and third categories, while richer solver diagnostics are needed for the first and fourth.

\section{Discussion and Limitations}

\method{} is most effective when the assembly can be decomposed into reusable engineering operations.  This is also its main limitation: new domains require new skills, resolvers, and validation checks.  The current benchmark emphasizes parametric mechanical and mold-oriented assemblies; free-form surfacing, complex organic shapes, tolerance-stack analysis, and downstream manufacturing simulation are outside the present scope.  The proxy comparisons are useful for debugging but should be replaced with official implementations or standardized benchmark submissions when available.  Finally, the current GRPO results show that solver feedback is necessary but not sufficient: a training signal must distinguish geometrically meaningful alternatives, not only executable syntax.

\section{Conclusion}

We introduced \method, a closed-loop framework for autonomous parametric \brep{} assembly modeling with LLM agents.  By separating LLM planning, deterministic operation-family resolution, and CAD-kernel execution, the framework converts open-loop script generation into solver-grounded skill trajectories.  Experiments on multi-category industrial assemblies show that the skill library and resolver can execute complete workflows, while learned policies achieve high valid rates but still face long-horizon exact-action errors.  The results suggest that future text-to-CAD systems should evaluate not only visual plausibility, but also solver execution, editability, and intermediate state validity.

\bibliographystyle{plainnat}
\bibliography{references}

\end{document}